\documentclass[runningheads]{llncs}
\usepackage[T1]{fontenc}
\usepackage{graphicx}
\usepackage{booktabs}
\usepackage[misc]{ifsym}
\usepackage[T1]{fontenc}
\usepackage{graphicx}
\usepackage{amsmath, amssymb}
\usepackage{booktabs, multirow, multicol}
\usepackage{hyperref}
\usepackage{xcolor}
\usepackage{siunitx}
\usepackage{enumitem}
\usepackage{wrapfig}
\usepackage{caption}
\usepackage{subcaption}

\usepackage[ruled, vlined, linesnumbered]{algorithm2e}

\def\name {TRAKNN}

\title{\name{}: Efficient Trajectory Aware Spatiotemporal kNN for Rare Meteorological Trajectory Detection}
\titlerunning{\name{} for Rare Meteorological Trajectory Detection}

\author{Guillaume Coulaud\inst{1}\ Davide Faranda\inst{2,3,4}}
\institute{University of Montpellier, Inria, CNRS, LIRMM, Montpellier, France \email{guillaume.coulaud@inria.fr} \and 
Laboratoire des Sciences du Climat et de l’Environnement, UMR 8212 CEA-CNRS-UVSQ, Université Paris-Saclay \& IPSL, CEA Saclay l’Orme des Merisiers, 91191 Gif-sur-Yvette, France \and
London Mathematical Laboratory, 8 Margravine Gardens, London W6 8RH, UK \and 
LMD/IPSL, ENS, Université PSL, École Polytechnique, Institut Polytechnique de Paris, Sorbonne Université, CNRS, Paris France\\
\email{davide.faranda@cea.fr}
}

\begin{document}
\maketitle

\begin{abstract}
Extreme weather events, such as windstorms and heatwaves, are driven by persistent atmospheric circulation patterns that evolve over several consecutive days. While traditional circulation-based studies often focus on instantaneous atmospheric states, capturing the temporal evolution, or trajectory, of these spatial fields is essential for characterizing rare and potentially impactful atmospheric behavior. However, performing an exhaustive similarity search on multi-decadal, continental-scale gridded datasets presents significant computational and memory challenges. In this paper, we propose TRAKNN (TRajectory Aware KNN), a fully unsupervised and data-agnostic framework for detecting geometrically rare short trajectories in spatio-temporal data with an exact kNN approach. TRAKNN leverages a recurrence-based algorithm that decouples computational complexity from trajectory length and efficient batch operations, maximizing computational intensity. These optimizations enable exhaustive analysis on standard workstations, either on CPU or on GPU.

We evaluate our approach on 75 years of daily European sea-level pressure data. Our results illustrate that rare trajectories identified by TRAKNN correspond to physically coherent atmospheric anomalies and align with independent extreme-event databases. 
\end{abstract}


\section{Introduction}

Extreme weather events such as storms, heatwaves or cold snaps have important impacts on ecosystems~\cite{ummenhoferExtremeWeatherClimate2017} and society~\cite{hoeppeTrendsWeatherRelated2016}. In a context of human-induced climate change, their frequency and intensity are evolving~\cite{stottHowClimateChange2016} and their socio-economic cost has increased over recent decades~\cite{forzieri2018escalating}, motivating sustained efforts to better understand their underlying physical drivers~\cite{hortonReviewRecentAdvances2016} and attribute the role of climate change~\cite{national2016attribution}. In Europe, many extreme events are associated with persistent large-scale atmospheric circulation patterns whose evolution unfolds over several consecutive days~\cite{hanleyRoleLargescaleAtmospheric2012}. While many existing circulation-based studies~\cite{horton2015contribution,jezequel2018role} focus on instantaneous atmospheric states (i.e., a spatial field at a single time step), the temporal evolution of spatial fields plays a central role in the development, amplification, and persistence of extremes. Capturing short spatio-temporal trajectories rather than isolated snapshots is therefore essential for characterizing rare atmospheric behavior. Detecting rare trajectories enables the systematic identification of unusual dynamical evolutions, supports the retrieval of historical analogues, and provides a principled basis for linking atmospheric dynamics to observed extreme impacts. Adopting a fully data-driven and unsupervised methodology further ensures that the approach remains problem-agnostic and data-agnostic, allowing systematic investigation without relying on predefined regimes, physical thresholds, or prior domain-specific knowledge.

From a data mining perspective, this setting naturally leads to anomaly detection in high-dimensional spatio-temporal data. Distance-based approaches such as Nearest-neighbor–based approaches~\cite{angiulli2002fast,ramaswamy2000efficient} are particularly attractive in this context due to their non-parametric nature, interpretability, and minimal modeling assumptions. However, applying such methods systematically to multi-decadal, continental-scale datasets presents substantial computational challenges. An exhaustive, exact comparison of trajectories requires quadratic scaling in the number of time steps and significant memory to load the data, which can quickly become prohibitive for datasets spanning several decades of daily observations on standard hardware.

In practice, these constraints often lead to subsampling, approximate similarity search, or reliance on specialized high-performance computing infrastructure. In contrast, we deliberately target a different design objective: enabling efficient, systematic, exhaustive trajectory-based analysis on standard hardware, without prohibitive memory requirements. Achieving this goal requires careful algorithmic design to eliminate redundant computations induced by overlapping temporal windows while preserving exactness.

Existing work on anomaly detection addresses aspects of this problem but does not directly tackle the combination of high spatial dimensionality, multi-decadal time horizons, and trajectory-based comparison at the continental scale under practical hardware constraints. This motivates the development of an efficient and memory-aware algorithm tailored to gridded spatio-temporal data.

This paper makes the following contributions:
\begin{itemize}
    \item We formulate a generic, fully unsupervised framework for detecting geometrically rare short trajectories in gridded spatio-temporal data, independent of the underlying physical variable.
    \item We derive an exact recurrence-based algorithm for trajectory distance computation, making the computational cost independent of the trajectory length.
    \item We propose an \name{} efficient algorithm that enables exhaustive multi-decadal analysis on standard workstations with a CPU and a CPU+GPU version.
    \item We illustrate our approach on 75 years of European daily sea-level pressure data, showing that rare trajectories exhibit coherent atmospheric anomalies and show agreement with independent extreme-event databases.
\end{itemize}

Together, these results illustrate that exact geometrically rare trajectories correspond to meaningful weather anomalies rarity, and the rarity estimation at continental and multi-decadal scales is computationally feasible and yields interpretable insights into extreme atmospheric behavior.

The remainder of the paper is organized as follows. We present the related work in Section~\ref{sec:related}. The problem and our algorithm are then described in Section~\ref{sec:method}. A performance evaluation of the algorithm scaling is presented in Section~\ref{sec:scaling}. We illustrate the method on a real-world case study in Section~\ref{sec:casestudy}
We discuss broader applications in Section~\ref{sec:discussion} and conclude in Section \ref{sec:conclusion}.

\section{Related Work}
\label{sec:related}
The detection of rare patterns in spatio-temporal datasets lies at the intersection of climate science, machine learning, and anomaly detection. In atmospheric sciences, extreme events are traditionally identified either through impact-based thresholds or circulation-based diagnostics. Many studies rely on flow-analogue approaches~\cite{jezequel2018role,ren2020attribution,collazo2024influence}, where analogues are defined as historical days whose large-scale fields are most similar to a target day under Euclidean distance. Such approaches focus primarily on instantaneous atmospheric states and do not explicitly account for the temporal evolution of circulation patterns. Building on this existing work on analogue and similarity search, we investigate the potential of a distance-based method to retrieve geometrically rare trajectories.

From a data mining perspective, rare trajectory detection can be formulated as unsupervised anomaly detection in high-dimensional time-series data. Classical distance-based methods such as $k$-nearest neighbors (\(k\)NN)~\cite{ramaswamy2000efficient,angiulli2002fast}, Local Outlier Factor ~\cite{breunigLOFIdentifyingDensity2000}, and Isolation Forest~\cite{liuIsolationForest2008} define anomalies through geometric isolation in feature space. These methods are attractive due to their minimal modeling assumptions and interpretability, but their application to multi-decadal spatio-temporal datasets is computationally challenging due to the cost of exhaustive similarity search.

Modern similarity search libraries such as FAISS~\cite{johnsonBillionScaleSimilaritySearch2021,douzeFAISSLIBRARY2025} enable large-scale nearest-neighbor retrieval, often using approximate indexing structures or dedicated computing hardware. However, they do not exploit the strong temporal overlap induced by sliding-window trajectory construction, resulting in redundant computations when applied directly to consecutive subsequences.

Time-series data mining has addressed related challenges in subsequence similarity search and motif discovery. The Matrix Profile (MP) framework~\cite{zhu2016matrix} introduced exact and scalable all-pairs similarity search for univariate time series using recurrence-based distance updates. Subsequent work extended this idea to multidimensional and streaming settings. Nevertheless, these approaches do not directly address the challenge of spatio-temporal data.

A key concern when applying distance-based methods in high dimensions is the phenomenon of distance concentration~\cite{beyerWhenNearestNeighbor1999,aggarwalSurprisingBehaviorDistance2001}. However, theoretical analyses have shown that nearest-neighbor methods remain meaningful when data lie on lower-dimensional structures or exhibit strong dependencies~\cite{durrantWhenNearestNeighbour2009}. Climate reanalysis fields are characterized by substantial spatial correlation and low intrinsic dimensionality, as evidenced by empirical orthogonal function analyses~\cite{hannachi2007empirical}. More generally, intrinsic dimensionality estimation methods~\cite{levinaMaximumLikelihoodEstimation2004} provide tools to quantify the effective degrees of freedom underlying high-dimensional observations.

Recent advances in deep learning have also been applied to spatio-temporal anomaly detection~\cite{wang2020deep,chalapathy2019deep}, including convolutional autoencoders, variational autoencoders, and recurrent neural networks. While these approaches can capture complex nonlinear structure, they typically require substantial training data, model tuning, and computational resources. Several work has investigate deep learning methods for the detection of extreme weather events~\cite{liu2016application,mudigonda2021deep,verma2023deep}.
In summary, existing approaches either (i) focus on instantaneous states, (ii) rely on approximate similarity search without accounting for the temporal overlap, or (iii) require complex parametric modeling. Our work instead targets exact, trajectory-aware, and computationally efficient $k$NN-based rarity detection tailored to gridded spatio-temporal datasets without strong hardware requirements. 

Additionally, the goal of our study is not to directly detect the extreme weather events, but to detect rare trajectories and assess 1) if they correspond to meaningful physical patterns, and 2) if they can be linked to extreme events.

\section{Methodology}
\label{sec:method}
\subsection{Problem Overview}
The core problem we address is how to systematically identify unusual atmospheric evolutions in large-scale gridded climate data. Consider a sequence of daily sea-level pressure maps over Europe spanning several decades. An extreme weather event, such as a windstorm or a heatwave, is rarely a single snapshot but rather a process that unfolds over multiple days. For example, an extratropical cyclone typically develops, matures, and decays over a period of three to seven days. During this evolution, the spatial pattern of pressure evolves in a coherent but potentially rare way.

Our goal is to detect such rare temporal evolutions without relying on predefined thresholds, labeled events, or domain-specific physical indices. We adopt a purely data-driven, geometric perspective: a sequence of consecutive daily maps of shape \(h\times w\) is treated as a \textit{trajectory} in a high-dimensional space. A trajectory is considered rare if it lies in a region of this space that is far from all other trajectories, i.e., if it is geometrically isolated. This isolation is quantified by the average distance to its $k$-nearest neighbors: trajectories with high average nearest-neighbor distance are considered isolated and therefore rare.

This formulation is intuitive: if an atmospheric evolution is unusual, it should have few close counterparts in the historical record. The challenge lies in making this idea computationally feasible for datasets containing tens of thousands of time steps and thousands of spatial grid points. The remainder of this section formalizes the problem and presents an efficient algorithm that makes exhaustive trajectory analysis practical on standard hardware.

\subsection{Problem Definition}
We address the task of identifying rare temporal evolutions within high-dimensional spatio-temporal datasets. Let $\mathbf{X} = \{X_1, X_2, \dots, X_n\}$ be a discrete-time sequence of spatial fields, where each field $X_t \in \mathbb{R}^{h \times w}$ is a \(h\times w\) matix of a physical or abstract variable (e.g., pressure, temperature, or pixel intensities) observed at time $t$.

We define a \textbf{spatio-temporal trajectory}  $T_t^{(d)}$ of duration $d$ starting at time $t$ as an ordered sequence of $d$ consecutive spatial fields:
\begin{equation}
T_t^{(d)} = (X_t, X_{t+1}, \dots, X_{t+d-1}) \in \mathbb{R}^{d \times h \times w}
\end{equation}
A trajectory is a third-order tensor representing a path in the high-dimensional configuration space of the system over the time interval $[t, t+d-1]$.

To quantify the similarity between two trajectories $T_i^{(d)}$ and $T_j^{(d)}$, we utilize the squared Euclidean distance $\mathcal{D}$ defined by the Frobenius norm of the difference of two trajectories. This captures the cumulative point-wise variance across both spatial and temporal dimensions:
\begin{equation}
\mathcal{D}(T_i^{(d)}, T_j^{(d)}) = \|T_i^{(d)}- T_j^{(d)}\|_F^2 = \sum_{m=0}^{d-1} \| X_{i+m} - X_{j+m} \|_F^2
\end{equation}

In many real-world systems, temporal persistence causes a trivial match effect, where trajectories overlapping in time appear highly similar. To ensure the discovery of non-trivial isolation, we define an \textbf{exclusion zone} $e$. Two trajectories $T_i$ and $T_j$ are only considered candidate neighbors if $|i - j| > e$. 

The objective is to compute a \textbf{rarity score} $s_t$, defined as the mean distance to the $k$-nearest neighbors (kNN) of $T_t^{(d)}$ within the set of all non-overlapping trajectories.
\begin{equation}
    \label{eq:rec}
    s_t = \frac{1}{k} \sum_{j \in \mathcal{N}_t} \mathcal{D}(T_t^{(d)}, T_j^{(d)}),
\end{equation}
where \(\mathcal{N}_t\) is the set of indices representing the k-nearest neighbors of \(T_t^{(d)}\).

\subsection{Trajectory Aware KNN}
We propose the \textbf{TRajectory Aware KNN} (\textbf{\name{}}) algorithm described in Algorithm~\ref{algo:algo}. It is divided into two parts: the efficient computation of all-pairs similarity between all spatial fields and the computation all the distance between trajectories with a recurrence-based optimization that decouples the computational complexity from the trajectory duration $d$.

\begin{algorithm}[ht]
\DontPrintSemicolon
\KwIn{Sequence of fields $\mathbf{X}\in \mathbb{R}^{h\times w \times n}$, duration $d$, exclusion zone $e$, neighbors $k$, batch size $b$}
\KwOut{Rarity scores $\mathbf{s} \in \mathbb{R}^{n-d+1}$}

\tcp{1. Symmetric Batched GeMM Spatial Distance Computation}
Precompute squared norms $N_i = \|X_i\|_F^2$ for $i=1 \dots n$\;
\For{batch index $i \leftarrow 1, b+1, \dots, n$}{
    \For{batch index $j \leftarrow i, i+b, \dots, n$}{
        $G \leftarrow B_i^\top B_j$ \tcp*{Compute only upper triangle}
        \For{local indices $u, v$ in batches}{
            $S_{u,v} \leftarrow N_u + N_v - 2G_{u,v}$\;
            $S_{v,u} \leftarrow S_{u,v}$ \tcp*{Exploit symmetry}
        }
    }
}

\tcp{2. Initialization and Recurrence}
\For{$j \leftarrow 1$ \KwTo $n-d+1$}{
    $D_{1,j} \leftarrow \sum_{m=0}^{d-1} S_{1+m, j+m}$\;
}
\For{$t \leftarrow 2$ \KwTo $n-d+1$}{
    \tcp{Vectorized update}
    $D_{t, \cdot} \leftarrow D_{t-1, \text{prev}} - S_{t-1, \text{prev}} + S_{t+d-1, \text{curr}}$\;
    
    \tcp{3. kNN Rarity Estimation}
    $\mathcal{N}_t \leftarrow \text{indices of } k \text{ smallest in } D_t \text{ where } |t-j| > e$\;
    $s_t \leftarrow \frac{1}{k} \sum_{j \in \mathcal{N}_t} D_{t,j}$\; 
}
\caption{\name{} algorithm}
\label{algo:algo}
\end{algorithm}

\paragraph{Optimized Spatial Distance Computation}
The \name{} algorithm first precomputes a spatial squared distance matrix $S \in \mathbb{R}^{n \times n}$. This repeated evaluation of pairwise spatial distances between high-resolution fields constitutes the dominant computational bottleneck. We reformulate the squared distance computation between two trajectories as a batch-wise General Matrix--Matrix Multiplication (GeMM) operation to maximize the computational intensity.

Let \(X_i, X_j \in \mathbb{R}^{h \times w}\) be two spatial fields. The squared Euclidean distance between them is computed as:
\begin{equation}
\|X_i - X_j\|_F^2 = \|X_i\|_F^2 + \|X_j\|_F^2 - 2 \langle X_i, X_j \rangle_F,
\end{equation}
where
\[
\langle X_i, X_j \rangle_F = \sum_{p=1}^{h}\sum_{q=1}^{w} (X_i)_{p,q}(X_j)_{p,q} \quad \text{ and }\quad \|X\|_F^2 = \langle X,X \rangle
\]

The norm of the spatial fields \(\|X_i\|_F^2\) is precomputed and cached, reducing subsequent distance evaluations to inner-product computations and vector additions. 
By reshaping each field \(X_i\) into a vector \(x_i \in \mathbb{R}^{m}\) with \(m = h \cdot w\), we aggregate several space fields into batches of dense matrices $B_1, B_2 \in \mathbb{R}^{m \times b}$. The batch of dot products is then obtained via a single Level-3 BLAS operation $G = B_1^{\top} B_2$. Since the spatial squared distance matrix is symmetric ($S_{i,j} = S_{j,i}$), we only compute $G$ for the upper triangular batches, effectively halving the number of required GeMM operations and maximizing FLOP utilization on modern hardware. This step corresponds to lines 1-7 of Algorithm~\ref{algo:algo}.

\paragraph{Constant-Time Trajectory Distance Recurrence}
Once \(S\) is computed, \name{} calculates trajectory distances. As two temporally consecutive trajectories $T_i^{(d)}$ and $T_{i-1}^{(d)}$ share $d-1$ spatial fields, the squared distance for any pair $(i, j)$ is updated in constant time $\mathcal{O}(1)$:
\begin{equation}
\label{eq:rec}
\mathcal{D}(T_{i}^{(d)}, T_{j}^{(d)}) = \mathcal{D}(T_{i-1}^{(d)}, T_{j-1}^{(d)}) - S_{i-1, j-1} + S_{i+d-1, j+d-1}
\end{equation}
This recurrence must be explicitly initialized in lines 8-9, then is incrementally computed row-wise, leveraging vectorized operations in line 11.

\paragraph{Complexity analysis} The computational complexity of \name{} is dominated by the initial spatial distance matrix calculation, which scales as $\mathcal{O}(hw \cdot n^2)$. The cost to compute all trajectories distance with Equation.~\ref{eq:rec} (line 6) is \(\mathcal{O}(n^2)\). The recurrence is initialized at a cost of \(\mathcal{O}(d\cdot n)\). The cost to retrieve the top \(k\)NN in line 11 is \(\mathcal{O}(n\log k)\). Therefore \name{} computational complexity is $\mathcal{O}(hw \cdot n^2)$. 
In terms of memory, the algorithm requires storing the $n \times n$ spatial distance matrix $S$, a vector of length \(n\) for the precomputed norms, and two vectors of length $(n-d+1)$, one for the trajectory distances and one for the scores. Since the algorithm also stores the spatio-temporal data of shape \(h\times w\times n\), the memory footprint is $\mathcal{O}(\max{(hw, n)\cdot n})$. 

\section{Performance Evaluation}
\label{sec:scaling}
\paragraph{Setup}
The code is written in Python 3.11 and Pytorch 2.5, allowing the algorithm to be executed seamlessly on both CPUs and GPUs. For the GPU version, only the batch-wise computation of the spatial matrix is done on the GPU, and the matrix is stored in the RAM. The code of the algorithms and all experiments is available on the following Github repository\footnote{https://github.com/GuillaumeCld/Trajectory-kNN}. The experiments are conducted on a laptop with an Intel CPU (i9-13900H@2.6GHz) with 32 GB of RAM and an NVIDIA GPU (RTX 2000 Ada, 8GB).

The execution time and memory usage of the algorithm are evaluated on synthetic random spatio-temporal data. As a baseline, we compare the computation time to the FAISS algorithm~\cite{douzeFAISSLIBRARY2025,johnsonBillionScaleSimilaritySearch2021}, a state-of-the-art library allowing similarity search on both CPUs and GPUs.

\paragraph{Execution time}

\begin{figure*}[ht!]
    \centering
    \begin{subfigure}[b]{.49\textwidth}
         \centering
         \includegraphics[width=\linewidth]{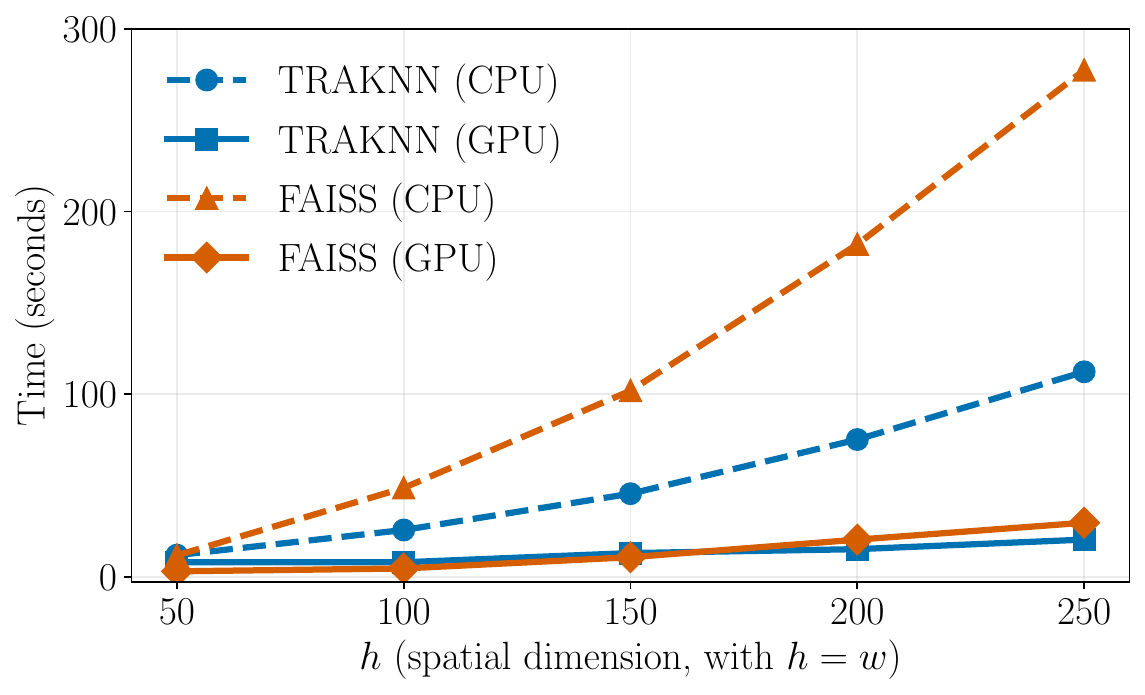}
         \caption{Varying spatial height \(h\) with \(w = h\)}
         \label{fig:scaling_spatial}
    \end{subfigure}
    \begin{subfigure}[b]{.49\textwidth}
         \centering
         \includegraphics[width=\linewidth]{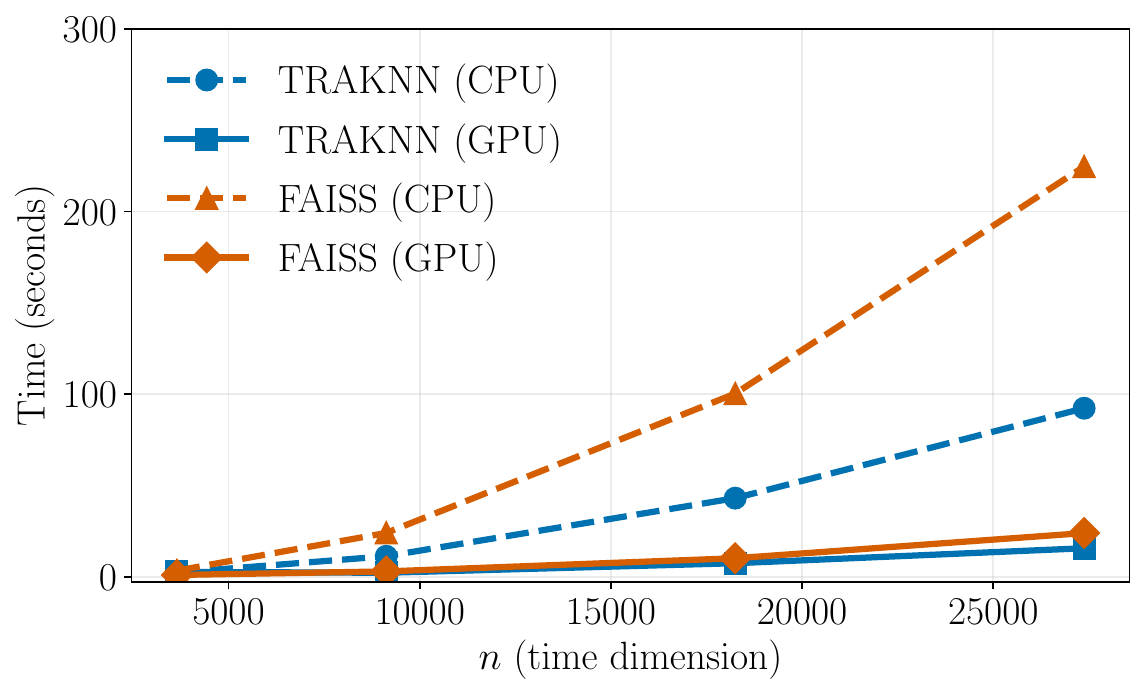}
         \caption{Varying timestamp \(n\) }
         \label{fig:scaling_time}
    \end{subfigure}
    \begin{subfigure}[b]{.49\textwidth}
         \centering
         \includegraphics[width=\linewidth]{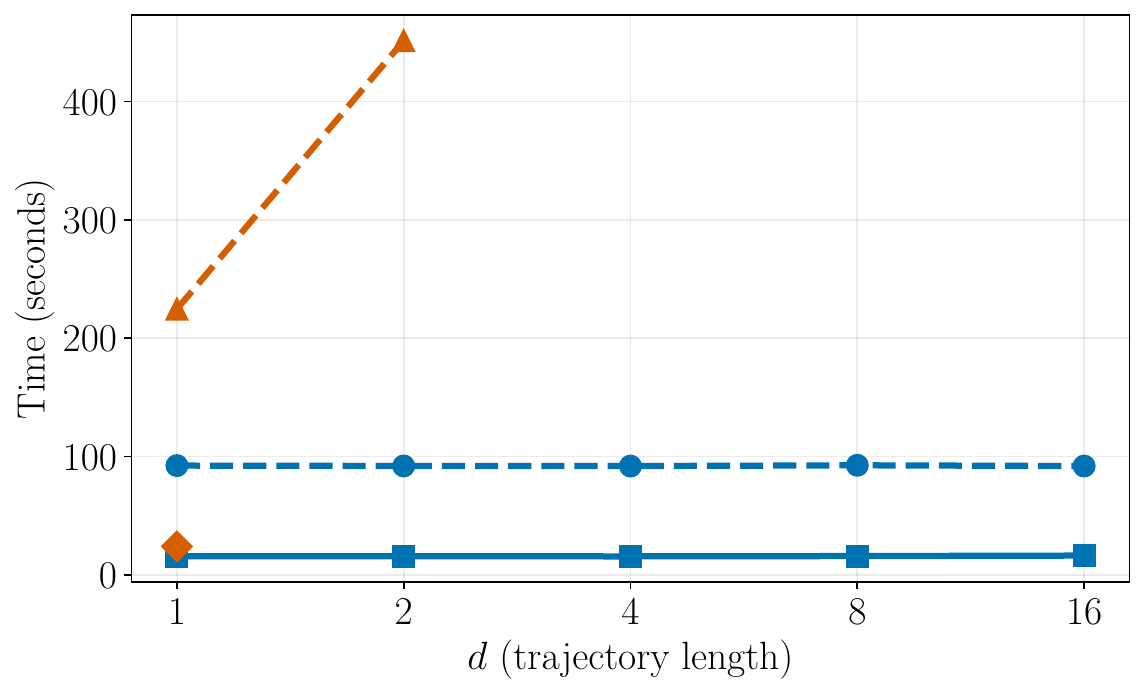}
         \caption{Varying trajectory length \(d\)}
         \label{fig:scaling_duration}
    \end{subfigure}
    \begin{subfigure}[b]{.49\textwidth}
         \centering
         \includegraphics[width=\linewidth]{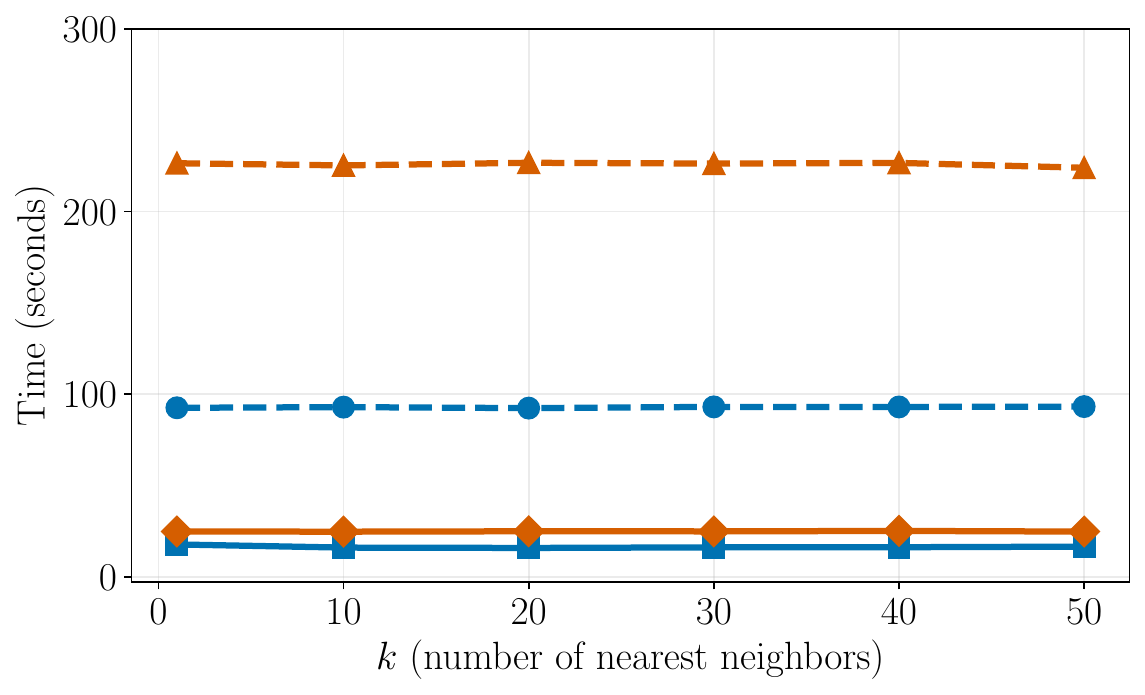}
         \caption{Varying number of neighbors \(k\)}
         \label{fig:scaling_k}
    \end{subfigure}
    \caption{Execution time (in seconds) of \name{} compared to FAISS on both CPU and GPU as a function of (a) spatial dimension length, (b) number of time steps, (c) trajectory length, and (d) number of neighbors. The default parameters are \(n=27,000\), \(h=180,w=280\), \(d=1\), and \(k=1\).}
    \label{fig:scaling}
\end{figure*}

Figure~\ref{fig:scaling} highlights the strong scalability of \name{} with respect to the spatial dimension  \(h\times w\)(Fig.~\ref{fig:scaling_spatial}), the number of time steps \(n\) (Fig.~\ref{fig:scaling_time}), the trajectory duration \(d\) (Fig.~\ref{fig:scaling_duration}), and the number of neighbors \(k\) (Fig.~\ref{fig:scaling_k}). For trajectories of length~1, \name{} achieves performance comparable to FAISS on GPU, while slightly outperforming it on CPU. More importantly, \name{} maintains nearly constant computation time as the trajectory length increases, demonstrating its efficiency for large-scale trajectory rarity analysis. In contrast, FAISS requires constructing and storing a database, which induces significant memory overhead and prevents it from scaling to longer trajectories.

\begin{figure}[ht]
    \centering
    \includegraphics[width=.45\linewidth]{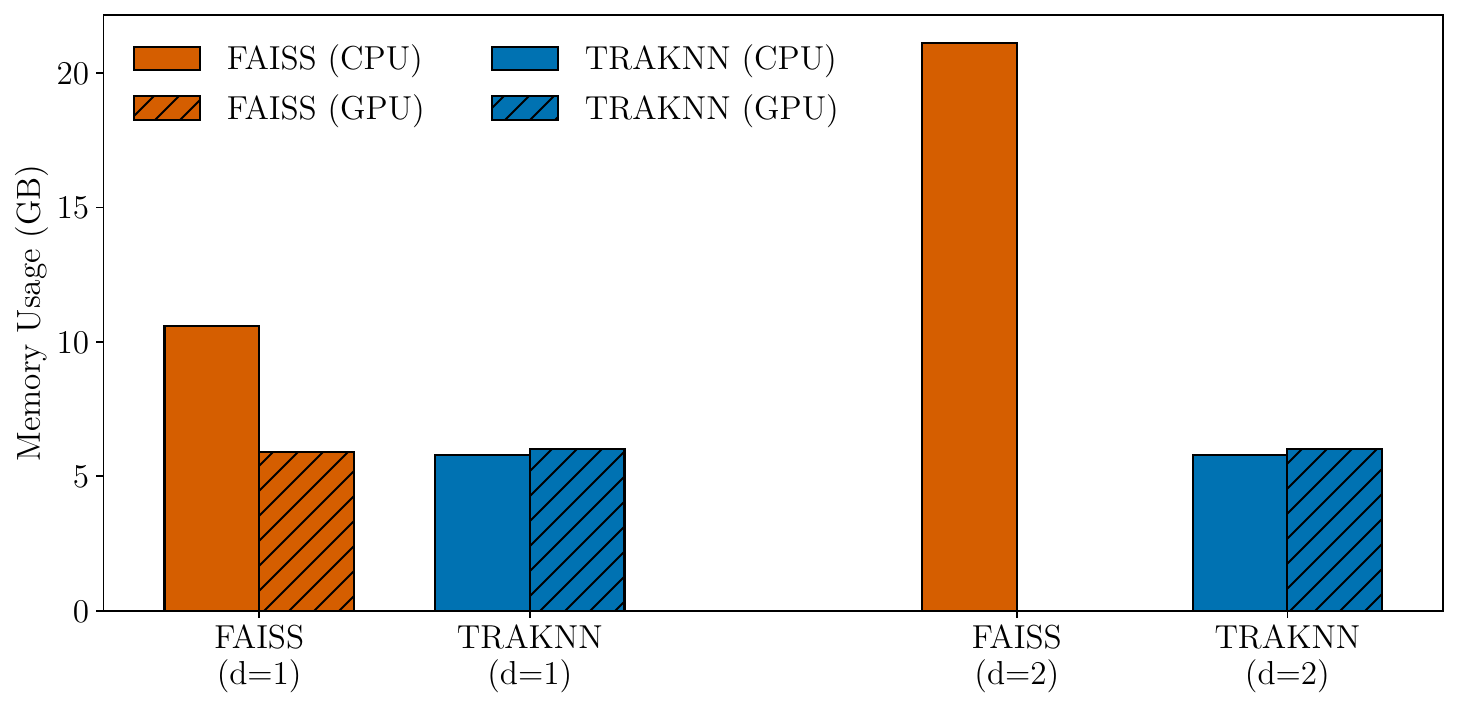}
    \caption{RAM peak memory usage per algorithm}
    \label{fig:memory}
\end{figure}

\paragraph{Memory usage} Figure~\ref{fig:memory} displays the peak RAM usage of all algorithms on the \(180\times 280\) grid with 75 years of daily observations corresponding to \(n\approx 27,000\), with \(k=10\), and \(d \in \{1, 5\}\).  It highlights that \name{} uses 6 GB of memory independently of \(d\), while for FAISS requires more memory, 10 GB for \(d=1\), and the memory usage is linear in \(d\). For \(d=2\) FAISS, needs more than 20 GB and cannot be used on a GPU.

\section{Case Study}
\label{sec:casestudy}
We illustrate the pertinence of the method on mean sea level-pressure (SLP) data over Europe (\(30^\circ\mathrm{N}-75^\circ\mathrm{N}\), \(30^\circ\mathrm{W}-40^\circ\mathrm{E}\)) from 1950 to 2025. The data corresponds to gridded maps with a resolution of \(0.25^\circ\) corresponding to images of resolution \(181\times 281\), i.e, approximately 50,000 spatial points, with daily observations over 75 years corresponding to approximately \(27,000\) time points. In total, we process more than a billion data points. The data is directly downloadable from the C3S Climate Data Store~\cite{hersbachERA5HourlyData2023}.

Our experiments are designed to answer the following three questions: 1) Is the Euclidean distance still meaningful for high-dimensional trajectories? 2) Do the geometrically rare trajectories correspond to meaningful atmospheric anomalies? 3) Can rare trajectories correspond to extreme weather events?

\subsection{Comparison with Dimension Reduction}\label{sec:pca}

\paragraph{Intrinsic dimension study} To better understand the geometric structure of the high-dimensional atmospheric trajectories, we estimate the intrinsic dimensionality (ID) of the trajectory space as a function of the trajectory duration $d$, i.e., the minimal number of variables needed to represent the data. Although each trajectory of length $d$ resides in an ambient space of dimension $d \times h \times w$, the effective degrees of freedom may be substantially lower due to strong spatial and temporal correlations. We therefore compute the averaged local ID using the maximum-likelihood $k$NN estimator~\cite{levinaMaximumLikelihoodEstimation2004}, which locally quantifies the ID. Given this estimator depends on the parameter \(k\) and \name{} allows to efficiently retrieves the \(k\) neighbors, we apply the estimator with \(k \in \{20, 30, 40\}\). To reduce the impact of the selection of \(k\), we report the macro averages and standard deviations of the local IDs.
This analysis serves two purposes: (i) to assess whether Euclidean nearest-neighbor distances remain meaningful despite the high ambient dimensionality, and (ii) to characterize how dynamical complexity evolves with increasing temporal embedding length. The results shown in Table~\ref{tab:dimension} reveal that the ID is two orders of magnitude lower than the ambient dimension, with a sub-linear growth and clear saturation of ID, increasing from $8 \pm 7$ for single-day fields to approximately $18 \pm 5$ for trajectories of length five days or longer. Notably, beyond five days, additional temporal context only increase slightly the ID, indicating that multi-day atmospheric evolutions remain confined to a structured low-dimensional manifold.

\begin{table}[]
    \centering
    \caption{Comparison of the ambient space dimension (\(d\times h\times w\)) and the mean local intrinsic dimension (ID) as a function of the trajectory duration on the mean sea level pressure}
    \label{tab:dimension}
    \begin{tabular}{c|cc}
    \toprule
       \(d\)  &  Ambient & ID \\\midrule
        1  &  50,000  & $8\pm 7$\\
        3  &  150,000 & $16\pm 6$\\
        5  &  250,000 & $18\pm 5$\\
        7  &  350,000 & $19\pm 4$\\
        15~ &  750,000 & $21\pm 3$\\
    \bottomrule
    \end{tabular}
\end{table}

\paragraph{Impact of PCA on detection} We now investigate if applying \name{} on the data with reduced spatial dimension impact the detection of the rarest trajectories. To reduce the spatial dimension, we use the PCA low rank approximation from PyTorch~\cite{halkoFindingStructureRandomness2011}, which performs a random SVD to approximate a lower rank \(q\) of the matrix. We set \(q=100\), then retain the components explaining 99\% of the variance. This allows to reduce the dimension from approximately 50,000 to 33. We apply \name{} with \(d=5\) and \(k=10\). Among the top 100 highest scores detected with and without dimension reduction, there are 99 common dates out of 100. The Spearman rank correlation between the common dates is 0.99, indicating almost the same rank ordering. Retaining 95\% of explained variance lower the number of dimension to 13, and results to 93 common dates, with a rank correlation of 0.96.

This result highlight with the previous result on the intrinsic dimension, highlights the pertinence of using the Euclidean distance to estimate the distance between trajectories despite the high dimension of the trajectories.

\subsection{Composite Analysis}\label{sec:composite}

To assess the physical coherence of the detected trajectories, we perform a composite analysis of the top 100 rarest atmospheric trajectories identified for a trajectory length of $d=5$ days with $k=10$. The objective is to examine whether high-rarity trajectories correspond to structured and interpretable large-scale circulation patterns, rather than isolated noise-driven configurations. To make the composite analysis, we use a standard procedure in atmospheric analysis and describe it in the next paragraph.

For each trajectory, we consider the anomaly fields relative to the long-term climatology and vectorize each spatial field into a state vector of dimension $h \times w$, corresponding to the number of grid points in the spatial domain. To account for the decreasing physical area of grid cells towards the poles, each grid point is weighted by the cosine of latitude. This area-weighting ensures that subsequent analysis reflects physically meaningful spatial structures and is not biased by higher grid density at high latitudes. The weighted anomaly vectors are then standardized by removing the mean and dividing by the standard deviation at each grid point. This step ensures that all spatial locations contribute comparably to the analysis, independent of their intrinsic variability. Then we reduce the dimensionality with the low rank PC and keep 95\% of the explained variance. We regroup similar dates with a clustering step performed in the reduced PCA space using the $k$-means algorithm with squared Euclidean distance. This procedure groups the rare trajectories into a small number of coherent dynamical regimes. For each cluster, composite anomaly maps are computed by averaging the original anomaly fields across all dates assigned to the cluster.

\begin{figure}
    \centering
    \includegraphics[width=.9\linewidth]{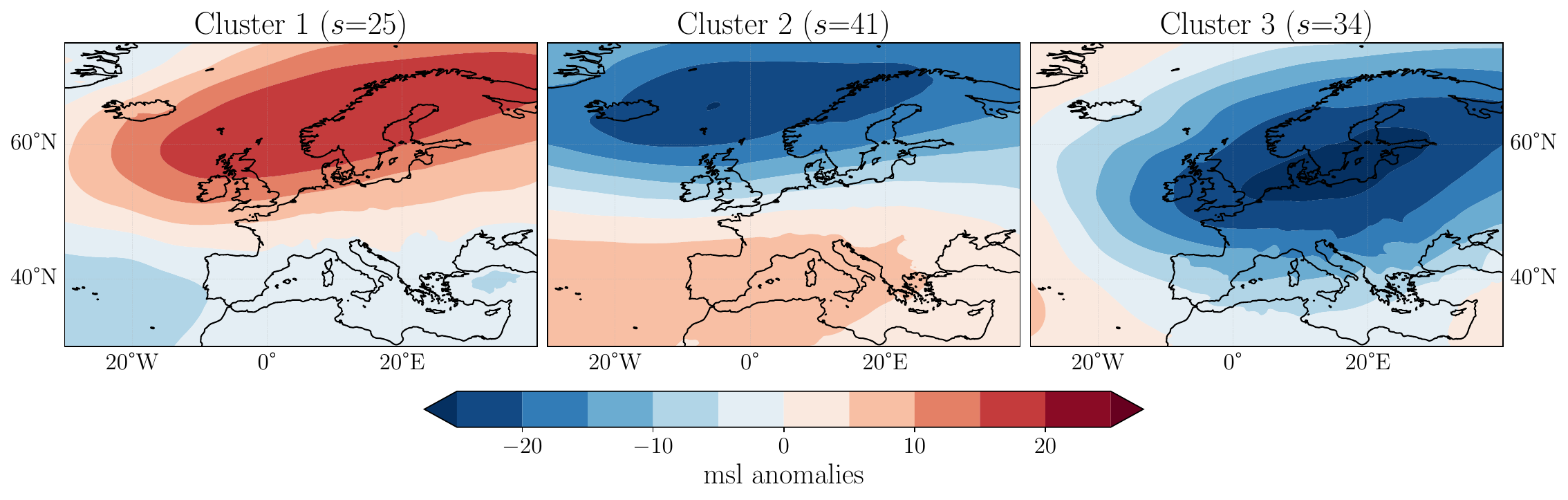}
    \caption{Composite analysis on the three main clusters of mean sea level (msl) pressure anomalies}
    \label{fig:composite}
\end{figure}

The resulting composites, displayed in Figure~\ref{fig:composite}, reveal persistent large-scale structures characterized by strong pressure gradients and spatially organized anomaly patterns. Especially, we find three relevant cluster of cluster size \(s\). The first cluster (\(s=25\)) displays high positive pressure anomalies over northern Europe and low negative anomalies on the south part of the domain.
The second cluster (\(s=41\)) is the opposite of the first, with high negative anomalies on the northern part and low positive anomalies in the lower part. The third cluster (\(s=34\)) correspond to a large low pressure-regime regime (negative anomalies) covering almost all Europe.

This analysis confirms that the highest-ranked rare trajectories correspond to dynamically coherent and physically interpretable circulation evolutions rather than random outliers.

\subsection{Matching to Extreme Events}\label{sec:matching}
We now investigate if the top 100 trajectories correspond to known extreme weather events, especially wind storms.

\paragraph{Databases description}
We use three publicly accessible and independent databases: XWS~\cite{robertsXWSOpenAccess2014}, CLIMK-WINDS~\cite{flynnCLIMKWINDSNew2025}, and EM-DAT~\cite{delforgeEMDATEmergencyEvents2025}. 

The XWS databases consists of the 50 most extremes winter European wind storms from 1979 to 2013 determined by a meteorological index based on the physical characteristics of the event. The analysis is conducted on the following domain \(35^\circ\mathrm{N}-70^\circ\mathrm{N}\), \(15^\circ\mathrm{W}-25^\circ\mathrm{E}\), among the 5730 detected storms, the 50 most extreme were retained for the catalog. The detected storms are split into two categories: the \textit{insurance storms}, which consisted of 23 storms that caused high insurance losses, and the \textit{large storms}, which comprise 27 storms selected as the top “non-insurance” storms based on their ranking in the storm severity index.

The CLIMK–WINDS database consists of the 50 most severe European winter wind storms from 1995 to 2015 detected over the following spatial domain  \(27^\circ\mathrm{N}-72^\circ\mathrm{N}\), \(22^\circ\mathrm{W}-45^\circ\mathrm{E}\) using a meteorological index.

The EM-DAT database is a global database of natural and technological disasters; specifically, it contains meteorological extreme events such as storms, heat waves, and cold snaps. For this work, we collect only the European extreme events and compute the rarity score on the full spatial domain from 1950 to 2025.

\paragraph{Matching Methodology} For fair comparison, we investigate the rare SLP trajectories on the same spatial domain and the same date range of each catalog. For each analysis, we retain the top 100 dates with the highest rarity score, an event is considered detected if it belongs to a rare trajectory, i.e., it belongs between the date and the date plus \(d\) days, corresponding to the trajectory duration. We always set \(k=10\) for the experiments.

\paragraph{Results}

Table~\ref{tab:storm_trajectory_length} summarizes the number of dates associated with storms for different trajectory lengths \(d \in \{1,3,5,7\}\) across the three databases: \textit{XWS}, \textit{CLIMK--WINDS}, and \textit{EM-DAT}. The results highlight how detection varies both with trajectory duration and with the nature of the database (meteorological vs.\ impact-based).

A general increase in the number of identified days is observed as the trajectory length \(d\) increases, particularly for the meteorological catalogs. 

For CLIMK--WINDS, detections steadily rise from 6 (1-day trajectories) to 12 (7-day trajectories) dates of the top 100 covering a storm. Similarly, for \textit{XWS} Large storms, the number increases from 3 (1-day) to 7 (7-day). This monotonic or near-monotonic growth suggests that longer trajectories better capture the temporal extent of severe winter wind storms. Since such systems typically evolve over several days, extending \(d\) increases the probability that at least one rare SLP configuration overlaps with a cataloged event. Nevertheless, increasing \(d\) reduces the number of days corresponding to insurance storms. The difference between \textit{Insurance} and \textit{Large} storms in XWS likely reflects the distinction between impact-based and purely meteorological severity rankings. Insurance storms may be more localized or associated with specific high-loss days, which explains their stronger representation at short durations.

In contrast, some categories exhibit non-monotonic behavior. For instance, EM-DAT general storms decrease from 8 detections at \(d=1\) to 5 at \(d=3\), then increase again for longer durations. This indicates that increasing trajectory duration does not uniformly improve detection across all datasets.

As a comparison, a random selection of 100 dates would find, on average over 100 runs, 0.7 CLIMK-WINDS storms and 0.4 XWS storms (large + insurance).

\begin{table}[ht]
\centering
\caption{Number of identified dates associated with a storm, as a function of the trajectory duration \(d\) (in days) with \name{}, by dataset and storm subtype, compared to an averaged random estimator.}
\label{tab:storm_trajectory_length}
\begin{tabular}{@{}ll*{5}{c}@{}}
    \toprule
    \textbf{Dataset} & \textbf{Storm type} & \multicolumn{4}{c}{\(d\)} & \textbf{Random} \\
    \cmidrule(lr){3-6}
     &  & \textbf{1} & \textbf{3} & \textbf{5} & \textbf{7} &  \\
    \midrule
    \multirow{1}{*}{CLIMK-WINDS~ }
        & -- & 6 & 9 & 11 & 12 & 0.7 \\    
    \midrule

    \addlinespace[0.5em]
    \multirow{2}{*}{XWS}
        & Insurance storms & 6 & 5 & 2 & 4 & 0.2 \\
        & Large storms     & 3 & 4 & 5 & 7 & 0.2 \\    
    \midrule

    \addlinespace[0.5em]
    \multirow{3}{*}{EM-DAT~ }
        & Storm (general)         & 8 & 5 & 7 & 4 & 0.9 \\
        & Storm (tropical)        & 0 & 0 & 1 & 1 & 0.1 \\
        & Storm (extratropical)~  & 3 & 2 & 3 & 3 & 0.4 \\
    \bottomrule
\end{tabular}
\end{table}

Additionally, when compared with the temperature extremes reported in EM-DAT, the instantaneous dimension (\(d=1\)) shows that, out of 100 dates, 9 coincide with severe winter conditions and 17 with cold waves. As \(d\) increases, the correspondence with temperature extremes decreases substantially. For instance, at \(d=5\), none of the 100 dates correspond to severe winter conditions, and only 7 correspond to cold waves.

\paragraph{Implications}

Overall, the results indicate that (i) longer trajectories improve detection for large-scale winter wind storms, particularly in meteorological catalogs; and (ii) detection is more consistent in catalogs constructed using meteorological severity indices (CLIMK--WINDS and XWS) than in multi-hazard disaster databases such as EM-DAT, (iii) spatial maps snapshots are more relevant for extreme temperature events

\section{Discussion}
\label{sec:discussion}
\paragraph{Practical Insight}

\name{} allows for the exhaustive analysis of multi-decadal, continental-scale datasets on standard workstations in a matter of minutes. This scalability enables practitioners to perform rapid ``what-if'' analyses, such as varying trajectory lengths or testing different variables, which was previously prohibitive due to the quadratic scaling of exact similarity searches. Furthermore, the method’s ability to handle gridded data directly for any trajectory length, without the need for dimensionality reduction or complex feature engineering, provides a transparent and accessible tool for historical analogue retrieval and extreme event attribution.

A key consideration in \(k\)NN-based algorithms is the choice of \(k\). One advantage of this approach is that the value of \(k\) has minimal influence on the computational cost. Moreover, the top \(k\) results obtained for a larger value of \(k\) naturally include the top \(k\) results for any smaller value. Consequently, the rarity score can be evaluated for multiple values of \(k\) with only minor post-processing overhead, avoiding the need to recompute the entire procedure from scratch. This enables users to efficiently assess how different values of \(k\) affect the score and the identification of the rarest trajectories in an ensemble-like manner.

While the memory footprint can appear substantial, in practical analogue-based atmospheric studies the available record typically spans at most \(\approx\)75 years of daily data, and the spatial domain is constrained by the physical scale of the phenomenon under investigation. Therefore, the problem size considered in our case study reflects realistic upper bounds for real-world applications size rather than an artificial limitation of the method. 

\paragraph{Relation to Matrix Profile} The proposed recurrence-based optimization is, in spirit, an extension of the MP formalism to spatio-temporal data. By utilizing the recurrence relation in Eq.~\ref{eq:rec}, we apply an update logic analogous to the MP efficient algorithms~\cite{zhu2016matrix,akbarinia2019efficient}, but adapted for sequences of any dimension rather than scalar values. Consequently, for the special case where $k=1$, our rarity score $s_t$ represents a \textit{Spatiotemporal Matrix Profile}, scaling the all-pairs similarity search philosophy to spatio-temporal fields and providing a robust foundation for identifying non-trivial motifs and anomalies in complex physical systems.

\section{Conclusion}
\label{sec:conclusion}
We proposed \name{} to address a gap in climate science by allowing to shift the analytical focus from instantaneous atmospheric snapshots to the temporal evolution of spatial fields of any duration. 
Future work, if required by real-world applications, could investigate out-of-core computation to account for large spatial or temporal domains that 1) might not fit in memory, and 2) lead to an important data loading bottleneck.
Additionally, future work will investigate the incorporation of trajectory-based analogues in the flow analogues analysis.

\section*{Acknowledgment}
We acknowledge funding by Région Occitanie, from the European Union’s Horizon 2020 research and innovation programme under grant agreement No. 101003469 (XAIDA), the ANR project TASE  PowDev (Strategic Development of the Power Grids of the Future, ANR-22-PETA-0016), the ANR project Templex (ANR-23-CE56-0002), and the ANR project TRACCS, Transformer la modélisation du climat pour les services climatiques (ANR-22-EXTR-0005, Extending).

\bibliographystyle{plain}
\bibliography{bibliography}

\begin{thebibliography}{10}

\bibitem{aggarwalSurprisingBehaviorDistance2001}
Charu~C. Aggarwal, Alexander Hinneburg, and Daniel~A. Keim.
\newblock On the {{Surprising Behavior}} of {{Distance Metrics}} in {{High
  Dimensional Space}}.
\newblock In Jan Van~den Bussche and Victor Vianu, editors, {\em Database
  {{Theory}} — {{ICDT}} 2001}, pages 420--434. Springer.

\bibitem{akbarinia2019efficient}
Reza Akbarinia and Bertrand Cloez.
\newblock Efficient matrix profile computation using different distance
  functions.
\newblock {\em arXiv preprint arXiv:1901.05708}, 2019.

\bibitem{angiulli2002fast}
Fabrizio Angiulli and Clara Pizzuti.
\newblock Fast outlier detection in high dimensional spaces.
\newblock In {\em European conference on principles of data mining and
  knowledge discovery}, pages 15--27. Springer, 2002.

\bibitem{beyerWhenNearestNeighbor1999}
Kevin Beyer, Jonathan Goldstein, Raghu Ramakrishnan, and Uri Shaft.
\newblock When {{Is}} “{{Nearest Neighbor}}” {{Meaningful}}?
\newblock In Catriel Beeri and Peter Buneman, editors, {\em Database {{Theory}}
  — {{ICDT}}’99}, pages 217--235. Springer.

\bibitem{breunigLOFIdentifyingDensity2000}
Markus~M. Breunig, Hans-Peter Kriegel, Raymond~T. Ng, and J{\"o}rg Sander.
\newblock Lof: Identifying density-based local outliers.
\newblock In {\em Proceedings of the ACM SIGMOD International Conference on
  Management of Data}, pages 93--104. ACM, 2000.

\bibitem{chalapathy2019deep}
Raghavendra Chalapathy and Sanjay Chawla.
\newblock Deep learning for anomaly detection: A survey.
\newblock {\em arXiv preprint arXiv:1901.03407}, 2019.

\bibitem{collazo2024influence}
Soledad Collazo, Solange Suli, Pablo~G Zaninelli, Ricardo Garc{\'\i}a-Herrera,
  David Barriopedro, and Jos{\'e}~M Garrido-Perez.
\newblock Influence of large-scale circulation and local feedbacks on extreme
  summer heat in argentina in 2022/23.
\newblock {\em Communications Earth \& Environment}, 5(1):231, 2024.

\bibitem{delforgeEMDATEmergencyEvents2025}
Damien Delforge, Valentin Wathelet, Regina Below, Cinzia Lanfredi~Sofia, Margo
  Tonnelier, Joris A.~F. van Loenhout, and Niko Speybroek.
\newblock The {{EM-DAT Emergency Events Database Archive}}.

\bibitem{douzeFAISSLIBRARY2025}
Matthijs Douze, Alexandr Guzhva, Chengqi Deng, Jeff Johnson, Gergely Szilvasy,
  Pierre-Emmanuel Mazaré, Maria Lomeli, Lucas Hosseini, and Hervé Jégou.
\newblock {{THE FAISS LIBRARY}}.
\newblock pages 1--17.

\bibitem{durrantWhenNearestNeighbour2009}
Robert~J. Durrant and Ata Kabán.
\newblock When is ‘nearest neighbour’ meaningful: {{A}} converse theorem
  and implications.
\newblock 25(4):385--397.

\bibitem{flynnCLIMKWINDSNew2025}
Clare~M. Flynn, Julia Moemken, Joaquim~G. Pinto, Michael~K. Schutte, and
  Gabriele Messori.
\newblock {{CLIMK}}–{{WINDS}}: A new database of extreme {{European}} winter
  windstorms.
\newblock 17(9):4431--4453.

\bibitem{forzieri2018escalating}
Giovanni Forzieri, Alessandra Bianchi, Filipe~Batista e~Silva, Mario A~Marin
  Herrera, Antoine Leblois, Carlo Lavalle, Jeroen~CJH Aerts, and Luc Feyen.
\newblock Escalating impacts of climate extremes on critical infrastructures in
  europe.
\newblock {\em Global environmental change}, 48:97--107, 2018.

\bibitem{halkoFindingStructureRandomness2011}
N.~Halko, P.~G. Martinsson, and J.~A. Tropp.
\newblock Finding {{Structure}} with {{Randomness}}: {{Probabilistic
  Algorithms}} for {{Constructing Approximate Matrix Decompositions}}.
\newblock 53(2):217--288.

\bibitem{hanleyRoleLargescaleAtmospheric2012}
John Hanley and Rodrigo Caballero.
\newblock The role of large-scale atmospheric flow and {{Rossby}} wave breaking
  in the evolution of extreme windstorms over {{Europe}}.
\newblock 39(21).

\bibitem{hannachi2007empirical}
Abdel Hannachi, Ian~T Jolliffe, David~B Stephenson, et~al.
\newblock Empirical orthogonal functions and related techniques in atmospheric
  science: A review.
\newblock {\em International journal of climatology}, 27(9):1119--1152, 2007.

\bibitem{hersbachERA5HourlyData2023}
Hans Hersbach, Bill Bell, Paul Berrisford, Gionata Biavati, Andr{\'a}s
  Hor{\'a}nyi, Joaqu{\'i}n Mu{\~n}oz~Sabater, Julien Nicolas, Carole Peubey,
  Raluca Radu, Iryna Rozum, Dinand Schepers, Adrian Simmons, Cornel Soci, Dick
  Dee, and Jean-No{\"e}l Th{\'e}paut.
\newblock {{ERA5}} hourly data on single levels from 1940 to present.
\newblock Copernicus Climate Change Service (C3S) Climate Data Store (CDS),
  2023.
\newblock doi: 10.24381/cds.adbb2d47 (Accessed on 10-06-2024).

\bibitem{hoeppeTrendsWeatherRelated2016}
Peter Hoeppe.
\newblock Trends in weather related disasters – {{Consequences}} for insurers
  and society.
\newblock 11:70--79.

\bibitem{horton2015contribution}
Daniel~E Horton, Nathaniel~C Johnson, Deepti Singh, Daniel~L Swain, Bala
  Rajaratnam, and Noah~S Diffenbaugh.
\newblock Contribution of changes in atmospheric circulation patterns to
  extreme temperature trends.
\newblock {\em Nature}, 522(7557):465--469, 2015.

\bibitem{hortonReviewRecentAdvances2016}
Radley~M. Horton, Justin~S. Mankin, Corey Lesk, Ethan Coffel, and Colin
  Raymond.
\newblock A {{Review}} of {{Recent Advances}} in {{Research}} on {{Extreme Heat
  Events}}.
\newblock 2(4):242--259.

\bibitem{jezequel2018role}
Agla{\'e} J{\'e}z{\'e}quel, Pascal Yiou, and Sabine Radanovics.
\newblock Role of circulation in european heatwaves using flow analogues.
\newblock {\em Climate dynamics}, 50(3):1145--1159, 2018.

\bibitem{johnsonBillionScaleSimilaritySearch2021}
Jeff Johnson, Matthijs Douze, and Hervé Jégou.
\newblock Billion-{{Scale Similarity Search}} with {{GPUs}}.
\newblock 7(3):535--547.

\bibitem{levinaMaximumLikelihoodEstimation2004}
Elizaveta Levina and Peter Bickel.
\newblock Maximum {{Likelihood Estimation}} of {{Intrinsic Dimension}}.
\newblock In {\em Advances in {{Neural Information Processing Systems}}},
  volume~17. MIT Press.

\bibitem{liuIsolationForest2008}
Fei~Tony Liu, Kai~Ming Ting, and Zhi-Hua Zhou.
\newblock Isolation forest.
\newblock In {\em Proceedings of the IEEE International Conference on Data
  Mining (ICDM)}, pages 413--422. IEEE, 2008.

\bibitem{liu2016application}
Yunjie Liu, Evan Racah, Joaquin Correa, Amir Khosrowshahi, David Lavers,
  Kenneth Kunkel, Michael Wehner, William Collins, et~al.
\newblock Application of deep convolutional neural networks for detecting
  extreme weather in climate datasets.
\newblock {\em arXiv preprint arXiv:1605.01156}, 2016.

\bibitem{mudigonda2021deep}
Mayur Mudigonda, Prabhat Ram, Karthik Kashinath, Evan Racah, Ankur Mahesh,
  Yunjie Liu, Christopher Beckham, Jim Biard, Thorsten Kurth, Sookyung Kim,
  et~al.
\newblock Deep learning for detecting extreme weather patterns.
\newblock {\em Deep Learning for the Earth Sciences: A Comprehensive Approach
  to Remote Sensing, Climate Science, and Geosciences}, pages 161--185, 2021.

\bibitem{national2016attribution}
National~Academies of~Sciences, Medicine, Division on~Earth, Life Studies,
  Board on~Atmospheric~Sciences, Committee on~Extreme Weather~Events, and
  Climate~Change Attribution.
\newblock {\em Attribution of extreme weather events in the context of climate
  change}.
\newblock National Academies Press, 2016.

\bibitem{ramaswamy2000efficient}
Sridhar Ramaswamy, Rajeev Rastogi, and Kyuseok Shim.
\newblock Efficient algorithms for mining outliers from large data sets.
\newblock In {\em Proceedings of the 2000 ACM SIGMOD international conference
  on Management of data}, pages 427--438, 2000.

\bibitem{ren2020attribution}
Liwen Ren, Tianjun Zhou, and Wenxia Zhang.
\newblock Attribution of the record-breaking heat event over northeast asia in
  summer 2018: the role of circulation.
\newblock {\em Environmental Research Letters}, 15(5):054018, 2020.

\bibitem{robertsXWSOpenAccess2014}
J.~F. Roberts, A.~J. Champion, L.~C. Dawkins, K.~I. Hodges, L.~C. Shaffrey,
  D.~B. Stephenson, M.~A. Stringer, H.~E. Thornton, and B.~D. Youngman.
\newblock The {{XWS}} open access catalogue of extreme {{European}} windstorms
  from 1979 to 2012.
\newblock 14(9):2487--2501.

\bibitem{stottHowClimateChange2016}
Peter Stott.
\newblock How climate change affects extreme weather events.
\newblock 352(6293):1517--1518.

\bibitem{ummenhoferExtremeWeatherClimate2017}
Caroline~C. Ummenhofer and Gerald~A. Meehl.
\newblock Extreme weather and climate events with ecological relevance: A
  review.
\newblock 372(1723):20160135.

\bibitem{verma2023deep}
Shikha Verma, Kuldeep Srivastava, Akhilesh Tiwari, and Shekhar Verma.
\newblock Deep learning techniques in extreme weather events: A review.
\newblock {\em arXiv preprint arXiv:2308.10995}, 2023.

\bibitem{wang2020deep}
Senzhang Wang, Jiannong Cao, and S~Yu Philip.
\newblock Deep learning for spatio-temporal data mining: A survey.
\newblock {\em IEEE transactions on knowledge and data engineering},
  34(8):3681--3700, 2020.

\bibitem{zhu2016matrix}
Yan Zhu, Zachary Zimmerman, Nader~Shakibay Senobari, Chin-Chia~Michael Yeh,
  Gareth Funning, Abdullah Mueen, Philip Brisk, and Eamonn Keogh.
\newblock Matrix profile ii: Exploiting a novel algorithm and gpus to break the
  one hundred million barrier for time series motifs and joins.
\newblock In {\em 2016 IEEE 16th international conference on data mining
  (ICDM)}, pages 739--748. Ieee, 2016.

\end{thebibliography}

\end{document}